\documentclass[11pt,a4paper]{article}
\usepackage[hyperref]{emnlp-ijcnlp-2019}
\usepackage{times}
\usepackage{latexsym}
\usepackage{stmaryrd}
\usepackage{amsmath,amssymb,amsfonts}
\usepackage{multicol,lipsum}
\usepackage{graphicx}
\usepackage{subcaption}

\usepackage{url}

\aclfinalcopy 

\title{Syntax-Aware Aspect Level Sentiment Classification with Graph Attention Networks}

\author{Binxuan Huang\\
  School of Computer Science\\
  Carnegie Mellon University\\
  {\tt binxuanh@cs.cmu.edu} \\\And
  Kathleen M. Carley \\
  School of Computer Science\\
  Carnegie Mellon University\\
  {\tt kathleen.carley@cs.cmu.edu} \\}

\date{}

\begin{document}
\maketitle
\begin{abstract}
Aspect level sentiment classiﬁcation aims to identify the sentiment expressed towards an aspect given a context sentence. Previous neural network based methods largely ignore the syntax structure in one sentence. In this paper, we propose a novel target-dependent graph attention network (TD-GAT) for aspect level sentiment classification, which explicitly utilizes the dependency relationship among words. Using the dependency graph, it propagates sentiment features directly from the syntactic context of an aspect target. In our experiments, we show our method outperforms multiple baselines with GloVe embeddings. We also demonstrate that using BERT representations further substantially boosts the performance.
\end{abstract}

\section{Introduction}
Aspect level sentiment classification aims to identify the sentiment polarity (eg. positive, negative, neutral) of an aspect target in its context sentence. Compared to sentence-level sentiment classification, which tries to detect the overall sentiment in a sentence, it is a more fine-grained task. Aspect level sentiment classification can distinguish sentiment polarity for multiple aspects in a sentence with various sentiment polarity, while sentence-level sentiment classification often fails in these conditions \cite{jiang2011target}. For example, in a sentence ``great food but the service was dreadful'', the sentiment polarity for aspects ``food'' and ``service'' are positive and negative respectively. In this case, however, it is hard to determine the overall sentiment since the sentence is mixed with positive and negative expressions.

Typically, researchers use machine learning algorithms to classify the sentiment of given aspects in sentences. Some early work manually designs features, eg. sentiment lexicons and linguistic features, to train classifiers for aspect level sentiment classification \cite{jiang2011target, wagner2014dcu}. Later,  various neural network-based methods became popular for this task \cite{tang2016aspect, wang2016attention}, as they do not require manual feature engineering. Most of them are based on long short-term memory (LSTM) neural networks \cite{tang2015effective, huang2018aspect} and few of them use convolutional neural networks (CNN) \cite{huang2018parameterized, xue2018aspect}. 

Most of these neural network based methods treat a sentence as a word sequence and embed aspect information into the sentence representation via various methods, eg. attention \cite{wang2016attention} and gate \cite{huang2018parameterized}. These methods largely ignore the syntactic structure of the sentence, which would be beneficial to identify sentiment features directly related to the aspect target. When an aspect term is separated away from its sentiment phrase, it is hard to find the associated sentiment words in a sequence. For example, in a sentence ``The food, though served with bad service, is actually great'', the word ``great'' is much closer to the aspect ``food'' in the dependency graph than in the word sequence. Using the dependency relationship is also helpful to resolve potential ambiguity in a word sequence. In a simple sentence ``Good food bad service'', ``good'' and ``bad'' can be used interchangeably. Using an attention-based method, it is hard to distinguish which word is associated with ``food'' or ``service'' among ``good'' and ``bad''. However, a human reader with good grammar knowledge can easily recognize that ``good'' is an adjectival modifier for ``food'' while ``bad'' is the modifier for ``service''.

In this paper, we propose a novel neural network framework named target-dependent graph attention network (TD-GAT), which leverages the syntax structure of a sentence for aspect level sentiment classification. Unlike these previous methods, our approach represents a sentence as a dependency graph instead of a word sequence. In the dependency graph, the aspect target and related words will be connected directly. We employ a multi-layer graph attention network to propagate sentiment features from important syntax neighbourhood words to the aspect target. We further incorporate an LSTM unit in TD-GAT to explicitly capture aspect related information across layers during recursive neighbourhood expansion. Though some work tries to incorporate syntax knowledge using recursive neural networks \cite{dong2014adaptive}, it has to convert the original dependency tree into a binary tree, which may move syntax related words away from the aspect term. Compared to \cite{dong2014adaptive}, one advantage of our approach is that it keeps the original syntax order unchanged.

We apply the proposed method to laptop and restaurant datasets from SemEval 2014 \cite{pontiki2014semeval}. Our experiments show that our approach outperforms multiple baselines with GloVe embeddings \cite{pennington2014glove}. We further demonstrate that using BERT representations \cite{devlin2018bert} boosts the performance a lot. In our analysis, we show that our model is lightweight in terms of model size. It achieves better performance and requires fewer computational resources and less running time than fine-tuning the original BERT model.

\section{Related Work}
Aspect level sentiment classification is a branch of sentiment analysis \cite{pang2008opinion}. The goal of this task is to identify the sentiment polarity of an aspect target within a given context sentence. 

Some early work first converts an extensive set of features, eg. sentiment lexicons and parse context, into feature vectors, then train a classifier like support vector machine (SVM) based on these feature vectors. \citet{wagner2014dcu} combine sentiment lexicons, distance to the aspect target, and dependency path distance to train an SVM classifier. \citet{kiritchenko2014nrc} also propose a similar method and they show that adding parse context features could improve the prediction accuracy by more than one percent.

Later, many neural network based methods are introduced to approach this aspect level sentiment classification task. A majority of the work uses LSTM neural networks to model the word sequence in a sentence. \citet{tang2015effective} use two LSTMs to model the left and right contexts of an aspect target, then they take two final hidden states as classification features. \citet{wang2016attention} introduce the attention mechanism into this task \cite{bahdanau2014neural}. They use the aspect term embedding to generate an attention vector to select important aspect-related words in a sentence. Following this work, \citet{huang2018aspect} use two LSTM networks to model sentences and aspects in a joint way and explicitly capture the interaction between aspects and context sentences. From the sentence aspect correlation matrix, they find important words in aspects as well as in sentences. \citet{li2018hierarchical} further improve these attention-based methods by incorporating position information.

Except for these LSTM-based methods, there are some other neural methods existing in the literature. \citet{tang2016aspect} propose a deep memory network which consists of multiple computation layers and each layer computes an attention vector over an external memory. There are also some attempts using convolutional neural networks (CNN) to approach this task \cite{huang2018parameterized,xue2018aspect}. Features generated from the aspect are used to control the information flow in the CNN applied to the sentence \cite{huang2018aspect}. Benefited from the rich linguistic knowledge learned from massive language modeling \cite{devlin2018bert}, researchers show great progress on this task using BERT representations \cite{sun2019utilizing}. \citet{xu2019bert} utilizes additional in-domain datasets to post-train BERT's weights and then fine-tune it on this task. However, such a method requires a large corpus for post-training and the fine-tuning also takes a lot of computation resources and time. 

Unlike previous discussed neural network-based methods, our approach explicitly utilizes the syntax structure among one sentence and these sentiment features are propagated towards the aspect target on the dependency graph instead of on the original word sequence. Some early work also tries to leverage the syntax structure \cite{dong2014adaptive,nguyen2015phrasernn}. They have to convert the original dependency tree into a binary tree first and place the aspect target at the root node. As a result, sentiment features can be propagated recursively from the leaf nodes to the root node. However, such conversion may move modifying sentiment words farther away from the aspect target, while our approach keeps the original syntax order unchanged.


\section{Method}
\subsection{Text Representation}
Given a sentence $s = [w_1 , w_2 , ..., w_i , .., $ $w_n ]$ with length n and an aspect target $w_i$, we first map each word into a low-dimensional word embedding vector. For each word $w_i$, we get one vector $x_i \in R^{d}$ where $d$ is the dimension of the word embedding space.

\begin{figure}[!h]
  \centering
  \includegraphics[width=0.8\linewidth]{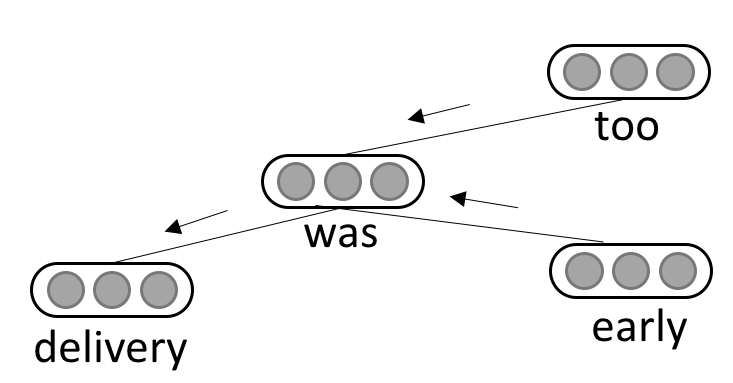}
  \caption{The dependency graph for ``delivery was early too''. Features can be propagated from neighbour nodes to the aspect node ``delivery''. }
\label{example}
\end{figure}

We transform the sentence into a dependency graph using an off-the-shelf dependency parser \cite{chen2014fast}. Each node represents a word and is associated with an embedding vector as its local feature vector. An undirected edge between two words means these two words are syntactically related. In Figure \ref{example}, we show an example of the dependency graph for a sentence ``delivery was early too''. For the target ``delivery'', we can propagate features from its 2-hop neighbourhood to 1-hop neighbourhood and then to itself.

For an aspect target with more than one word, we first replace the whole target word sequence with a special symbol ``\_\_target\_\_'', then pass the modified sentence into the dependency parser. As a result, there is a meta-node representing the target sequence in the dependency graph and its local feature vector is the average of embedding vectors of words in the target.

\subsection{Graph Attention Network}
A graph attention network (GAT) \cite{velivckovic2017graph} is a variant of graph neural network \cite{scarselli2009graph} and is a key element in our method. It propagates features from an aspect's syntax context to the aspect node. Given a dependency graph with $N$ nodes, where each node is associated with a local word embedding vector $x$, one GAT layer compute node representations by aggregating neighbourhood's hidden states. With an $L$-layer GAT network, features from $L$ hops away can be propagated to the aspect target node.

Specifically, given a node $i$ with a hidden state $h_l^i$ at layer $l$ and the node's neighbours $n[i]$ as well as their hidden states, a GAT updates the node's hidden state at layer $l+1$ using multi-head attentions \cite{vaswani2017attention}. The update process is as follows

\begin{align}
    h_{l+1}^i &=\bigparallel_{k=1}^K \sigma(\sum_{j\in n[i]} \alpha_{lk}^{ij}W_{lk}h_l^{j}) \label{eq1}\\
    \alpha_{lk}^{ij} &= \frac{exp(f(a_{lk}^T[W_{lk}h_l^i||W_{lk}h_l^j]))}{\sum_{u\in n[i]}exp(f(a_{lk}^T[W_{lk}h_l^i||W_{lk}h_l^u]))}
\end{align}

where $\bigparallel$ represents vector concatenation, $\alpha_{lk}^{ij}$ is the attention coefficient of node $i$ to its neighbour $j$ in attention head $k$ at layer $l$. $W_{lk} \in R^{\frac{D}{K}\times D}$ is a linear transformation matrix for input states. $D$ is the dimension of hidden states. $\sigma$ denotes a sigmoid function. $f(\cdot)$ is a LeakyReLU non-linear function \cite{maas2013rectifier}. $a_{lk} \in R^{\frac{2D}{K}}$ is an attention context vector learned during training. 

For simplicity, we can write such feature propagation process as 
\begin{align}
    H_{l+1} = GAT(H_{l},A;\Theta_l)
\end{align}
where $H_l\in R^{N\times D}$ is the stacked states for all nodes at layer $l$, $A\in R^{N\times N}$ is the graph adjacent matrix. $\Theta_l$ is the parameter set of the GAT at layer $l$. 

\subsection{Target-Dependent Graph Attention Network}
To utilize the target information in such a GAT network explicitly, we further use an LSTM to model the dependency for the aspect target across layers, which is also helpful for overcoming noisy information in a graph \cite{huang2019inductive}. 
The basic idea is that at layer $0$ the hidden state for an aspect target node $h_0^t$ is only dependent on the target's local features and at each layer $l$ information related with the target from $l$-hop neighbourhood is added into the hidden state by the LSTM unit.

Given the previous hidden state $h_{l-1}^t$ and cell state $c_{l-1}^t$ for any target node $t$, we first get a temporary hidden state $\hat{h_l^t}$ by aggregating its neighbour information using equation \ref{eq1}. Then we take this temporary hidden state as a new observation for an LSTM unit and update the hidden state at layer $l$ as follows:

\begin{align}
i_l &= \sigma(W_i  \hat{h_l^t}+U_i  h_{l-1}+b_i) \\
f_l &= \sigma(W_f \hat{h_l^t}+ U_f  h_{l-1}+b_f) \\
o_l &= \sigma(W_o  \hat{h_l^t}+U_o  h_{l-1}+b_o)\\
\hat c_l &=  tanh(W_c  \hat{h_l^t}+U_c  h_{l-1}+b_c)\\
c_l &= f_l \circ c_{l-1}+i_l\circ \hat c_l\\
h_l &= o_l \circ tanh(c_l)
\end{align}

where $\sigma(\cdot)$ and $tanh(\cdot)$ are the sigmoid function and hyperbolic tangent function respectively. $W_i, U_i, W_f$, $U_f, W_o,U_o, W_c,U_c$ are parameter matrices and $b_i,b_f,b_o,b_c$ are bias vectors to be learned during training. Symbol $\circ$ represents element-wise multiplication. $i_t$, $f_t$ and $o_t$ are input gate, forget gate and output gate, which control the information flow. 

In summary, the feed-forward process of our target-dependent graph neural network can be written as

\begin{align*}
    H_{l+1}, C_{l+1} &= LSTM(GAT(H_l,A;\Theta_l),(H_l,C_l))\\
    H_0,C_0 &= LSTM(XW_p+[b_p]_N, (0,0))
\end{align*}
 
where $C_l$ is the stacked cell states of the LSTM at layer $l$. The initial hidden state and cell state of the LSTM are set as 0. $W_p\in R^{d\times D}$ is the projection matrix that transforms stacked embedding vectors $X$ into the dimension of hidden states and $[b_p]_N$ represents stacking the bias vector $b_p$ N times and forms a bias matrix with dimension $R^{N\times D}$. Similarly, we can also replace the LSTM unit with a GRU unit to model the layer-wise dependency for the target.

\subsection{Final Classification}
With $L$ layers of our TD-GAT networks, we get a final representation for our aspect target node. We just retrieve the corresponding hidden state $h^t_L$ for the aspect target node from all the node representations $H_L$.

We map the hidden state $h^t_L$ into the classification space by a linear transformation. Afterwards, the probability of a sentiment class $c$ is computed by a softmax function:
\begin{align}
    P(y=c) = \frac{exp(Wh^t_L+b)_c}{\sum_{i\in C}exp(Wh^t_L+b)_i}
\end{align}
where $W,b$ are the weight matrix and bias for the linear transformation, $C$ is the set of sentiment classes.

The final predicted sentiment polarity of an aspect target is the label with the highest probability. We minimize the cross-entropy loss with $L_2$ regularization to train our model
\begin{align*}
loss = -\sum_{c\in C} I(y=c)\cdot log(P(y=c))+\lambda ||\Theta||^2
\end{align*}

\noindent where $I(\cdot)$ is an indicator function. $\lambda$ is the $L_2$ regularization parameter and $\Theta$ is the set of all the parameters in our model.

\section{Experiments}
\subsection{Datasets}
We adopt two widely used datasets from SemEval 2014 Task 4 \cite{pontiki2014semeval} to validate the effectiveness of our method. These are two domain-specific datasets collected from laptop and restaurant reviews. Each data point is a pair of a sentence and an aspect term. Experienced annotators tagged each pair with sentiment polarity. Following recent work \cite{tay2017learning, huang2018parameterized}, we first take 500 training instances as development set\footnote{The splits can be found at \href{https://github.com/vanzytay/ABSA_DevSplits}{https://github.com/vanzytay/ABSA\_DevSplits}.} to tune our model. We then combine the development dataset and training dataset to train our final model. Statistics of these two datasets are given in Table 1.

\begin{table}[!h]
\centering
\label{data}
\begin{tabular}{|l|l|l|l|}
\hline
Dataset          & Positive & Neutral & Negative \\ \hline
Laptop-Train     & 767      &   373   & 673      \\ \hline
Laptop-Dev     & 220      & 87     & 193      \\ \hline
Laptop-Test      & 341      & 169     & 128      \\ \hline
Restaurant-Train & 1886     & 531     & 685      \\ \hline
Restaurant-Dev & 278     & 102     & 120      \\ \hline
Restaurant-Test  & 728      & 196     & 196      \\ \hline
\end{tabular}
\caption{Statistics of the datasets. }
\end{table}

\subsection{Implementation Details}
 We use the Stanford neural parser \cite{chen2014fast} to get dependency graphs. We try two embedding methods in this paper. One is 300-dimensional GloVe embeddings \cite{pennington2014glove}, where we just retrieve the corresponding embedding vector for each token in graphs. Another is BERT representations \cite{devlin2018bert}, where we use the large uncased English model with dimension 1024 implemented in PyTorch \footnote{https://github.com/huggingface/pytorch-pretrained-BERT}. The input of the BERT model is a text pair formatted as ``[CLS]'' + sentence + ``[SEP]'' + aspect
+ ``[SEP]''. The representations of the sentence are used for the downstream aspect-level sentiment classification task. Because the tokenizers used in the parser and BERT are different, we get the BERT representations for tokens in dependency graphs by averaging the corresponding representations of sub-word units (``wordpiece'') from BERT. For example, the representation of the token ``overload'' is the average of representations of two sub-words ``over'' and ``\#\#load''. Once the word representations are initialized, they are fixed during training. 

We set the dimension of hidden states as 300 in our experiments. For the BERT representations, we first map word representations into 300 dimensional vectors by a linear projection layer. We use 6 attention heads in our model. We train our model with batch size of $32$. We apply $l_2$ regularization with term $\lambda$ $10^{-4}$ and dropout \cite{srivastava2014dropout} on the input word embedding with rate $0.7$. We first use Adam \cite{kingma2014adam} optimizer with learning rate $10^{-3}$ to train our model, then switch to stochastic gradient descent to fine-tune and stabilize our model. 
 
 We implemented our model using PyTorch Geometric \cite{fey2019fast} on a Linux machine with Titan XP GPUs.

\subsection{Baseline Comparisons}
To validate the effectiveness of our method, we compare it to following baseline methods:

\textbf{Feature-based SVM} utilizes n-gram features, parse features and lexicon features for aspect-level sentiment classification \cite{kiritchenko2014nrc}

\textbf{TD-LSTM} is a direct competitor against our method. It uses two LSTM networks to model the preceding and following contexts surrounding the aspect term, while we use GAT to model the syntax context around an aspect. The last hidden states of these two LSTM networks are concatenated for predicting the sentiment polarity \cite{tang2015effective}.

\textbf{AT-LSTM} first models the sentence via a LSTM model. Then it combines the hidden states from the LSTM with the aspect term embedding to generate the attention vector. The final sentence representation is the weighted sum of the hidden states \cite{wang2016attention}.

\textbf{MemNet} applies attention multiple times on the word embeddings, and the output of last attention is fed to softmax for prediction \cite{tang2016aspect}.

\textbf{IAN} uses two LSTM networks to model the sentence and aspect term respectively. It uses the hidden states from the sentence to generate an attention vector for the target, and vice versa. Based on these two attention vectors, it outputs a sentence representation and a target representation for classification \cite{ma2017interactive}. 

\textbf{PG-CNN} is a CNN based model where aspect features are used as gates to control the feature extraction on sentences \cite{huang2018parameterized}.

\textbf{AOA-LSTM} introduces an attention-over-attention (AOA) based network to model aspects and sentences in a joint way and explicitly capture the interaction between aspects and context sentences \cite{huang2018aspect}.

\textbf{BERT-AVG} uses the average of the sentence representations to train a linear classifier.

\textbf{BERT-CLS} is a model where we directly use the representation of ``[CLS]'' as a classification feature to fine-tune the BERT model for paired sentence classification. We fine-tune it for 5 epochs using Adam optimizer with batch size 8 and learning rate $10^{-5}$.

\begin{table}[!h]
\begin{tabular}{lcc}
\hline \hline
                 & Laptop & Restaurant \\ \hline
Feature+SVM      & 70.5   & 80.2       \\ 
TD-LSTM          & 68.1   & 75.6       \\ 
AT-LSTM          & 68.9   & 76.2       \\ 
MemNet           & 72.4   & 80.3       \\ 
IAN              & 72.1   & 78.6       \\ 
PG-CNN           & 69.1   & 78.9       \\ 
AOA-LSTM         & 72.6   & 79.7       \\  \hline
TD-GAT-GloVe (3) & {73.7}   & 81.1       \\ 
TD-GAT-GloVe (4) & \textbf{74.0}   & 80.6       \\ 
TD-GAT-GloVe (5) & 73.4   & \textbf{81.2}       \\ \hline \hline
BERT-AVG            & 76.5   & 78.7       \\ 
BERT-CLS            & 77.1   & 81.2       \\\hline
TD-GAT-BERT (3)  & 79.3   & 82.9       \\ 
TD-GAT-BERT (4)  & 79.8   & \textbf{83.0}       \\ 
TD-GAT-BERT (5)  & \textbf{80.1}   & 82.8       \\ \hline \hline
\end{tabular}
\caption{Comparison results of different methods on laptop and restaurant datasets. Numbers in parentheses indicate number of layers in our model.}
\label{compare}
\end{table}

The comparison results are shown in Table \ref{compare}. With GloVe embeddings, our approach TD-GAT-GloVe (k), where k is the number of layers, outperforms all these previous methods. Among these baselines, Feature-based SVM achieves strong performance on this task, which indicates the importance of feature engineering and syntax knowledge. 

 As one direct competitor, TD-LSTM propagates sentiment features from the beginning and the end of the sentence to the aspect target, while our model propagates features from syntax dependent words to the target on a dependency graph. Compared to TD-LSTM, our model shows superior performance, which directly proves the necessity of incorporating syntax information.
 
 Using BERT representations further boosts the performance of our model. BERT-AVG, which uses BERT representations without fine-tuning, achieves surprisingly excellent performance on this task. After fine-tuning, the performance of BERT-CLS becomes even better. However, we observe that such fine-tuning is quite unstable. The model cannot converge in some trials.

 Even though the original BERT model already provides strong prediction power, our model consistently improves over BERT-AVG and BERT-CLS, which indicates that our model can better utilize these semantic representations. The accuracy of our model reaches about 80\% and 83\% on the laptop and restaurant datasets respectively.

\subsection{Effects of Target Information}
In this section, we provide an ablation study to show the effects of explicitly capturing target information. In the ablated model, we remove the LSTM unit in our TD-GAT model, so that it cannot utilize the aspect target information explicitly. We denote this ablated model as GAT.

As shown in Table \ref{ablation}, explicitly capturing aspect target information consistently improves the performance of the TD-GAT-GloVe over the GAT-GloVe model. On average, the accuracy of TD-GAT-GloVe increased by 1.2 percentage points. Capturing aspect-related information explicitly across layer is also useful for the BERT-based model as well. Even though the target information has been embedded in the BERT representation because of the contextual language modeling, TD-GAT-BERT still outperforms the GAT-BERT model. On average, the explicit target information contributes 0.95 percentage points to the final performance of the TD-GAT-BERT.

\begin{table}[!h]
\resizebox{0.5\textwidth}{!}{
\begin{tabular}{lcccccc}
\hline \hline
Dataset      & \multicolumn{3}{c}{Laptop} & \multicolumn{3}{c}{Restaurant} \\ \hline 
layer        & 3       & 4       & 5       & 3         & 4        & 5        \\ \hline
GAT-GloVe    & 73.0    & 72.1    & 72.4    & 79.6      & 80.0     & 79.7     \\ 
TD-GAT-GloVe & 73.7    & 74.0    & 73.4    & 81.1      & 80.6     & 81.2     \\ \hline
GAT-BERT     & 78.1    & 78.5    & 78.5    & 82.6      & 82.2     & 82.3     \\ 
TD-GAT-BERT  & 79.3    & 79.8    & 80.1    & 82.9      & 83.0     & 82.8     \\ \hline\hline
\end{tabular}
}
\caption{An ablation study shows the effect of explicit target information.}
\label{ablation}
\end{table}

\subsection{Effects of Model Depth}

\begin{figure}[!h]
\centering
  \includegraphics[width=\linewidth]{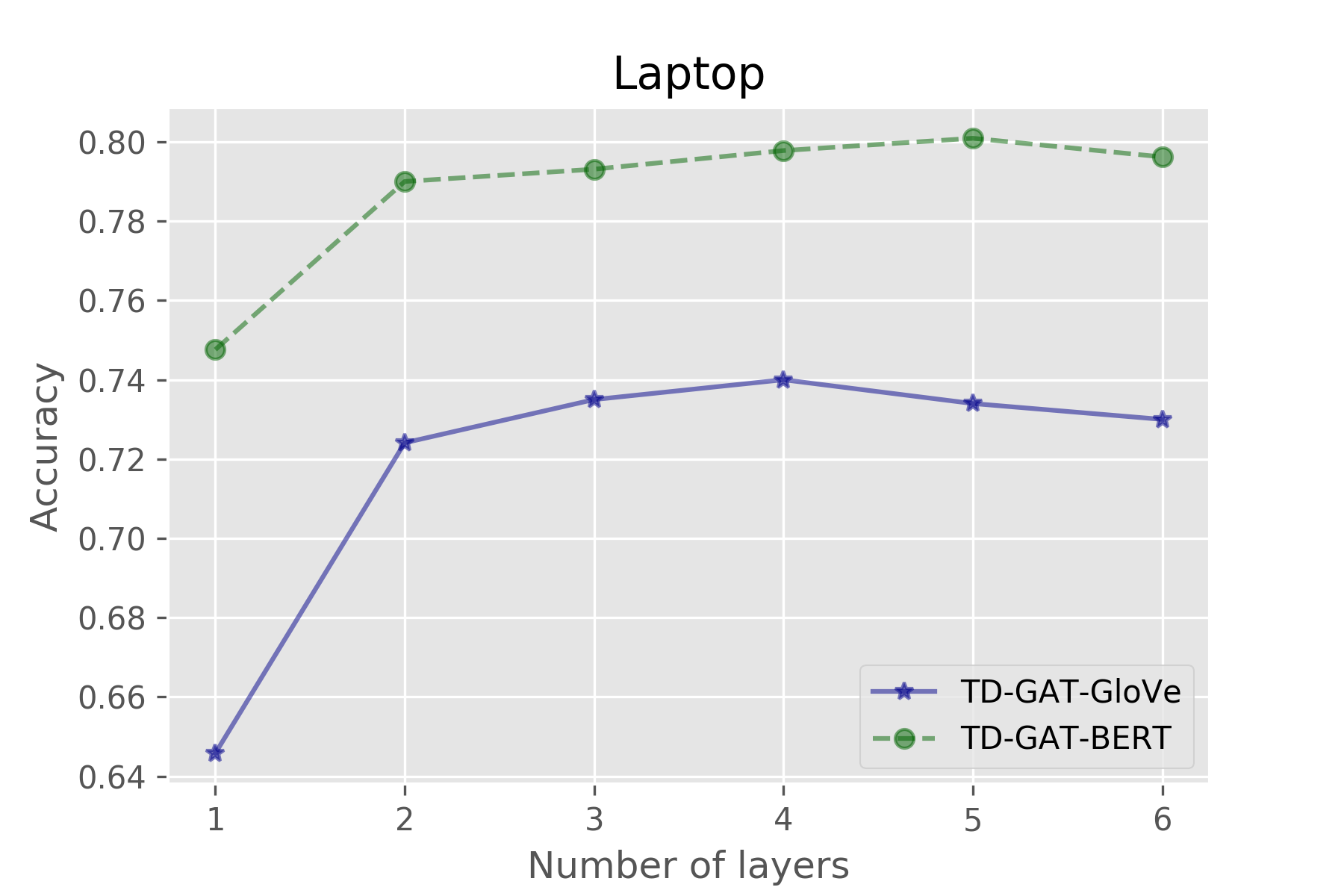}
  \includegraphics[width=\linewidth]{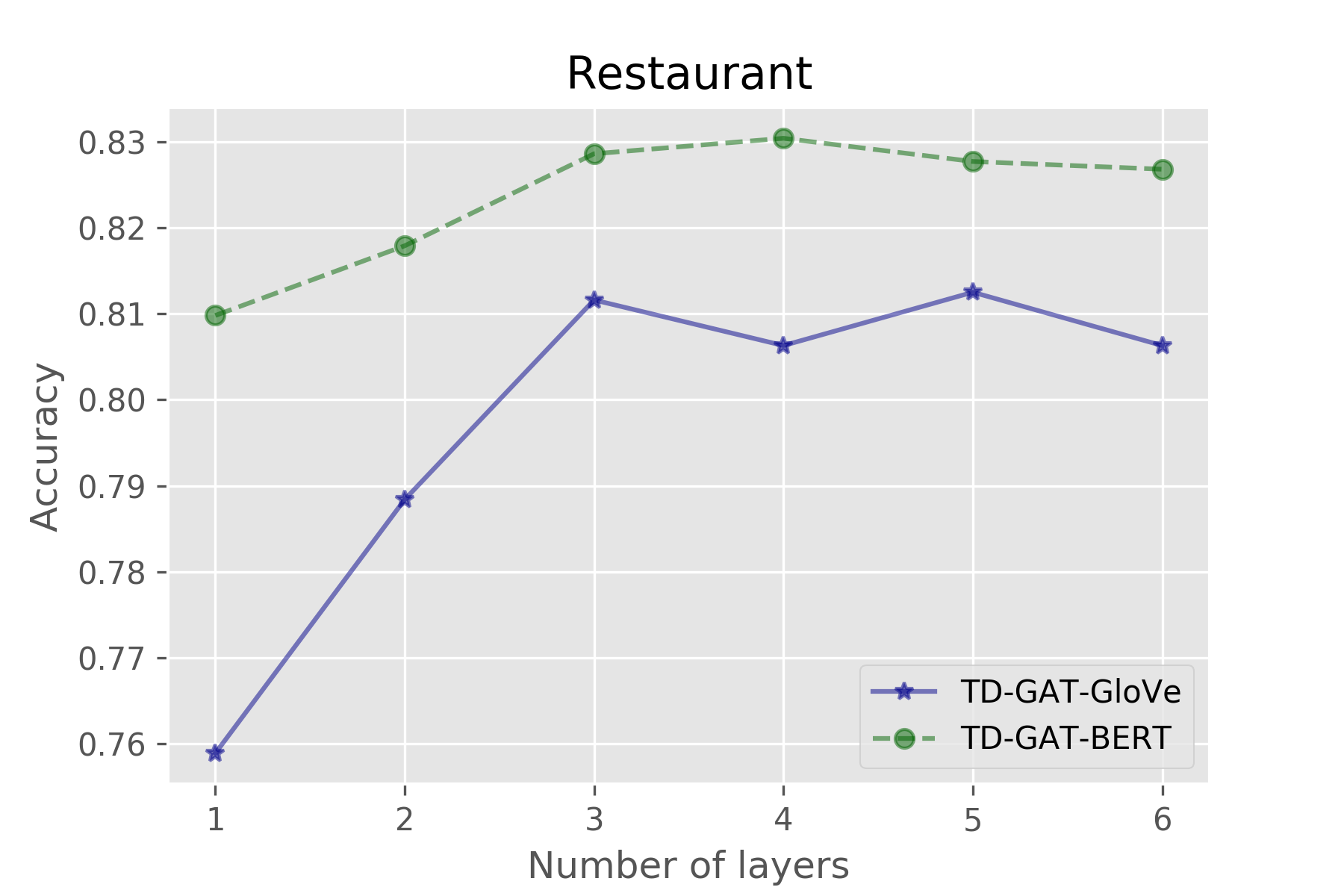}
\caption{The impact of model depth (number of layers).}
\label{depth}
\end{figure}

We explore the impact of model depth (number of layers) in this section. For our TD-GAT model, we vary its model depth ranging from 1 to 6. As shown in Figure \ref{depth}, a one-layer TD-GAT model with GloVe embeddings does not work well, which implies target-related sentiment words are usually 2-hops away from the aspect target. Increasing the model depth to 3 would greatly improve the performance of the TD-GAT-GloVe model.

Unlike the TD-GAT-GloVe model, our model with BERT representations are more robust to the model depth. Even a one-layer TD-GAT-BERT model still achieves satisfactory results on both datasets. One possible reason is that BERT representations already incorporate context words into these semantic representations. Hence nodes at one-hop away may consist of some global information. However, increasing the model depth still improve the performance in this case and our model reaches its optimal performance when model depth is larger than 3.

\subsection{Model Size}
We compare the model size of our TD-GAT model to various baseline methods as well as the BERT model. For these baseline methods, we use an open source PyTorch implementation \footnote{https://github.com/songyouwei/ABSA-PyTorch} to measure their model sizes. 

\begin{table}[!h]
\centering
\begin{tabular}{lc}
\hline\hline
Models           & Model size ($\times 10^6$)  \\ \hline
TD-LSTM          & 1.45      \\ 
MemNet (3)       & \textbf{0.36}      \\ 
IAN              & 2.17        \\ 
AOA-LSTM         & 2.89        \\ \hline
TD-GAT-GloVe (3) & 1.00        \\ 
TD-GAT-GloVe (4) & 1.09        \\ 
TD-GAT-GloVe (5) & 1.18       \\ \hline\hline
BERT-CLS         & 335.14      \\ \hline
TD-GAT-BERT (3)  & \textbf{1.30}    \\ 
TD-GAT-BERT (4)  & 1.39     \\ 
TD-GAT-BERT (5)  & 1.49      \\ \hline\hline
\end{tabular}
\caption{The model size (number of parameters) of our model as well as baselines.}
\label{size}
\end{table}

The sizes of all these models are given in Table \ref{size}. Using the same dimension of hidden states, our TD-GAT-GloVe has a lower model size compared to these LSTM-based methods. MemNet is the model ranks the first in terms of the model size. The size of TD-GAT-BERT increases by $0.3\times10^6$ because of the linear projection layer applied on the input word representations. When we switch from GloVe embeddings to BERT representations, the training time for a three-layer TD-GAT model on the restaurant dataset only increases from 1.12 seconds/epoch to 1.15 seconds/epoch. On the contrary, fine-tuning the BERT model takes about 226.50 seconds for each training epoch. Training our TD-GAT-BERT model requires much fewer computation resources and less time compared to fine-tuning the original BERT model.

\section{Conclusion}

In this paper, we propose a novel target-dependent graph attention neural network for aspect level sentiment classification. It leverages the syntactic dependency structure of a sentence and uses the syntax context of an aspect target for classification. Compared to those methods applied on word sequences, our approach places modifying sentiment words closer to the aspect target and can resolve potential syntactic ambiguity. In our experiments, we demonstrate the effectiveness of our method on laptop and restaurant datasets from SemEval 2014. Using GloVe embeddings, our approach TD-GAT-GloVe outperforms various baseline models. After switching to BERT representations, we show that TD-GAT-BERT achieves much better performance. It is lightweight and requires fewer computational resources and less training time than fine-tuning the original BERT model.

To the best of our knowledge, this paper is the first attempt directly using the original dependency graph without converting its structure for aspect level sentiment classification. Many potential improvements could be made in this direction. In this work, the local feature vector of an aspect node is the average of embedding vectors of words in the aspect and each word in the aspect is equally important. Future work could consider using an attention mechanism to focus on important words in the aspect. Since this work only uses the dependency graph and ignores various types of relations in the graph, we plan to incorporate dependency relation types into our model and take part-of-speech tagging into consideration as well in the future. We would also like to combine such a graph-based model with a sequence-based model to avoid potential noise from dependency parsing errors.

\section*{Acknowledgments}
This work was supported in part by the Office of Naval Research (ONR) N000141812108 and N000141712675. The views and conclusions contained in this document are those of the authors and should not be interpreted as representing the official policies, either expressed or implied, of the ONR.

\bibliography{reference}
\bibliographystyle{acl_natbib}

\end{document}